\documentclass[conference]{IEEEtran}
\IEEEoverridecommandlockouts
\usepackage{cite}
\usepackage{amsmath,amssymb,amsfonts}
\usepackage{algorithmic}
\usepackage{graphicx}
\usepackage{textcomp}
\usepackage{xcolor}
\usepackage{amsmath}
\usepackage[cmintegrals]{newtxmath}
\usepackage{bm}
\usepackage{array}
\usepackage{url}
\usepackage{enumitem}
\usepackage{multirow} 
\usepackage{multicol}
\usepackage[colorlinks=true, urlcolor=blue, citecolor=blue, linkcolor=darkgray]{hyperref}

\def\BibTeX{{\rm B\kern-.05em{\sc i\kern-.025em b}\kern-.08em
    T\kern-.1667em\lower.7ex\hbox{E}\kern-.125emX}}

\begin{document}

\title{Robust Readout of Cross-Point Resistive Arrays \\ via Time-Domain Optimization
\thanks{ This work was supported by JST CREST grant number JPMJCR2555, JSPS KAKENHI Grant Number JP22H05002 and 26KJ1229, Japan }
}

\author{
Taro Tako$^{1,*}$, Hiroshi Kaminaga$^{2}$, and Taisuke Kobayashi$^{1}$
\thanks{$^{1}$T. Tako and T. Kobayashi are with the National Institute of Informatics (NII) and with The Graduate University for Advanced Studies (SOKENDAI), 2-1-2 Hitotsubashi, Chiyoda-ku, Tokyo, 101-8430, Japan}%
\thanks{$^{2}$ H. Kaminaga is with the CNRS-AIST JRL (Joint Robotics Laboratory), IRL 3218, National Institute of Advanced Industrial Science and Technology (AIST), 1-1-1 Umezono, Tsukuba, Ibaraki 305-8568, Japan}%
\thanks{$^{*}$ Corresponding author: tako@nii.ac.jp}
}

\maketitle

\begin{abstract}
Cross-point resistive arrays are attractive for high-density tactile sensing because they reduce wiring complexity. However, conventional frequency-based readout methods can become inaccurate in robotic sensing environments, where long interconnects and measurement and environmental noise degrade the observed signals. We propose a time-domain optimization method for robust readout of cross-point resistive arrays. The method determines resistance values by fitting a time-domain observation model to the measured waveform over the interval, rather than relying solely on spectral amplitudes. By exploiting the full waveform, the proposed method improves robustness under noisy observation conditions. In simulations of a 16×16 array, the proposed method reduced the mean readout error by 20\% compared with the baseline method. Furthermore, validation over 1,000 trials showed that the proposed method achieves more robust estimation than the conventional method, reducing the mean error and standard deviation by 24.6\% and 77.6\%, respectively. The improvement was also confirmed in cross-point array simulations with resistance distributions emulating tactile sensor responses.
\end{abstract}

\begin{IEEEkeywords}
Resistive sensor array, AC measurement, robotics sensing
\end{IEEEkeywords}

\section{Introduction}
\begin{figure}[tb]
    \centering
    \includegraphics[width=0.9\linewidth]{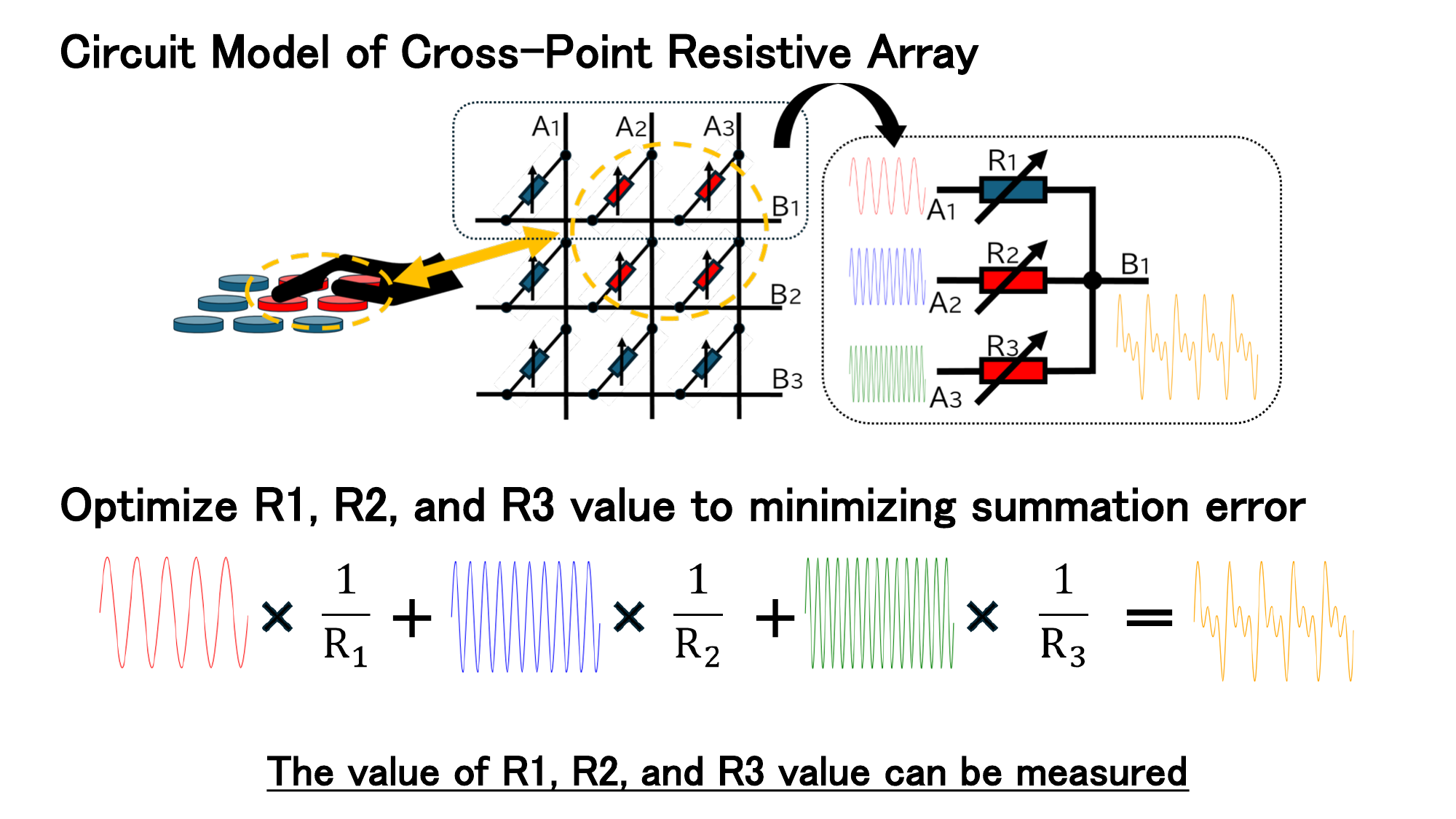}
    \caption{Conceptual overview of the proposed readout method}
    \label{fig:overview}
\end{figure}

Robot-embedded tactile sensors require large-area and dense coverage for accurate environmental perception and full-body coverage of the robot~\cite{cheng2019comprehensive, ohmura2006conformable, mukai2008development}.
Cross-point resistive arrays are widely used to realize large-area sensing in robots~\cite{Kagami2004,Kanazawa2020,tako2024tactile}.
In general, a cross-point resistive array is configured using switches for each row and column, and the readout target resistor is determined by the switch configuration.
Since the number of switches increases linearly with the number of resistors, the time required to read all resistances also increases, along with the readout error caused by the parasitic resistance of the switches.
Many previous studies on cross-point resistive arrays have focused on reducing readout errors~\cite{Kagami2004,shu2015,shiiki2023,dominguez2024fast,ghouchani2024improved}.
Alternating-current (AC) measurements~\cite{hu2019high,tong2023synchronous,li2024ultrafast} are promising because they eliminate switches in the circuit and enable simultaneous readout of multiple sensors, thereby avoiding timing mismatches among sensors and allowing simultaneous readout independent of array size.
Existing AC measurement methods for cross-point arrays distinguish element responses by assigning different frequency components to different channels and recovering the corresponding responses using spectral analysis.
However, these methods rely on frequency-domain information to recover individual sensor responses, and their accuracy can degrade significantly in the presence of noise and limited sampling conditions.
To address this limitation, we propose a time-domain optimization method, as illustrated in Fig.~\ref{fig:overview}.
The proposed method determines them from time-domain signals. 
This improves robustness to noise and enables accurate measurement even under partially degraded signal conditions.
We evaluate the proposed method in simulation using two circuit types and three conditions, including resistance distributions designed to emulate tactile sensor responses, and demonstrate its effectiveness.

\section{Principle of the AC measurement}
In AC measurement, sensing elements in a cross-point array are modeled as branches sharing a common terminal. 
A sinusoidal AC voltage applied to the $j$th channel at time $t$ is defined as follows:
\begin{align}
  \label{formula:input_signal}
  V_j(t) = A_j\sin (\omega_j t + \phi_j)
\end{align}
where $A_j$, $\omega_j$, and $\phi_j$ denote the amplitude, angular frequency, and phase of the input signal, respectively.
To read out the responses of different elements simultaneously, different frequencies are assigned to different channels.
If $M$ elements are measured simultaneously, $M$ different frequencies are required.
Since the channels share a common terminal, the measured current is the sum of all the branch currents through the resistors ${R_j}$:
\begin{align}
  \label{formula:sum_R_current}
  I_{\mathrm{o}}(t) = \sum_{j=1}^{M} {V_j(t)}/{R_j}
\end{align}

The measured current $I_{\mathrm{o}}(t)$ contains the frequency components of all applied signals. 
Because the input frequencies are known in advance, the response of each element can be identified from the corresponding component in the measured composite current. 
The spectrum of $I_{\mathrm{o}}(t)$ is obtained from the Fourier transform, $\hat{I}(f)$ is the amplitude spectrum, and $f$ is the frequency.
Under ideal conditions, a purely resistive element does not introduce additional frequency distortion.
The resistance value can be obtained from the amplitude of the corresponding spectral component.
The amplitude of the applied voltage is $A_j$ from eq.~\ref{formula:input_signal} and the resistance value can be calculated as follows:
\begin{align}
    \label{formula:ideal_estimate_R}
    \hat{R}_j = {A_j}/{\hat{I}(f_j)}.
\end{align}

Therefore, simultaneous measurement of multiple resistive elements is possible, as long as each element is associated with a distinct input frequency.
In a cross-point array, unique frequencies are assigned to the columns, and the response of each element is identified from the corresponding spectral component in the measured row current. 
The response of the element $R(i,j)$ is obtained as the amplitude of the frequency assigned to column $j$, extracted from the spectrum of the measured row current $I_{\mathrm{o},i}(t)$ by Fourier transform.
As a result, all elements in the array can be read out simultaneously.

In principle, the number of readable channels can be increased by assigning different frequencies to different elements. 
In practice, however, reliable separation of individual responses requires sufficient frequency spacing, which limits the number of usable channels within a given bandwidth. 
As the channel count increases, a wider frequency band is required, while accurate recovery of spectral amplitudes becomes more difficult under noise, limited observation windows, and signal degradation along long interconnects. 
These constraints are particularly relevant to robotic tactile sensing, where flexible sensor placement and long wiring are often unavoidable. 
Therefore, AC measurement methods that rely primarily on frequency-domain separation may suffer from reduced robustness as the channel count increases, reducing the practical advantage of cross-point arrays for high-density tactile sensing.

\section{Proposed Method}
To reduce the dependence on frequency-domain separation, the proposed method directly models the observed current waveform in the time domain and then derives the corresponding resistance values.
To improve measurement accuracy, we employ an observation model that accounts for the behavior of the resistive elements as well as non-ideal measurement effects.
For simplicity, the following formulation considers a single-row cross-point structure with $M$ columns. 
The current corresponding to the $j$th resistor is modeled as
\begin{align}
\label{formula:obs_current_R}
  I_j(t) = \alpha_j \sin(\Omega_j t + \Psi_j),
\end{align}
where $\alpha_j$, $\Omega_j$, and $\Psi_j$ denote the current amplitude, angular frequency, and phase, respectively. 
The phase reflects the non-ideal circuit effects such as parasitic inductance and capacitance in the circuit, while the effective angular frequency may deviate slightly under non-ideal sampling conditions.

We formulate an optimization problem for a single-row cross-point structure with $M$  columns. 
The observed current is expressed as eq.~\ref{formula:sum_R_current}.
We minimize the mismatch between the modeled and measured current waveforms.
Let $I_m(t)$ denote the measured current waveform.
In practice, the current is sampled at discrete time steps, and the objective function is defined over the observation samples as
\begin{align}
  \label{formula:obj_func}
    E = \frac{1}{2} \sum_{k=0}^{T} \left( I_o[k] - I_m[k] \right)^2,
\end{align}
where $I_m[k]$ is the measured current at the $k$th sample. 
By minimizing $E$, we obtain the model parameters, from which the amplitudes $\alpha_j$ are then used to determine the resistor values in the corresponding columns.
The resistance values are determined from Eq.~\ref{formula:ideal_estimate_R}. 
Moreover, the proposed method can be extended to larger cross-point arrays in a manner similar to conventional AC measurement.

The optimization is formulated as a nonlinear least-squares problem, whose convergence depends on the initial values. 
The computational cost of each optimization step increases with both the number of signal components and the number of samples.
Compared to the AC measurement method using the Fast Fourier Transformation (FFT), the proposed method requires iterative optimization for each observed row and therefore incurs additional computational overhead.
However, since both methods are applied row-wise, their total cost increases  similarly with array size. 
Thus, the main trade-off is between computational simplicity and measurement robustness.

\begin{table*}[tb]
    \centering
    \caption{Result for circuit~\ref{item:circuit1}, consisting of 16 parallel resistors, under condition~\ref{item:condition1}, where each resistor is $10~\mathrm{k}\Omega$.}
    \label{tab:result_of_exp1_circuit1}
    \scalebox{0.6}[1]{
      \begin{tabular}{c|cccccccccccccccc}
          \hline
          Frequency \lbrack~Hz~\rbrack
          & 5k & 6k & 7k & 8k & 9k & 10k & 11k & 12k & 13k & 14k & 15k & 16k & 17k & 18k & 19k & 20k \\ \hline \hline
            FFT(w/o noise) \lbrack~\%~\rbrack
            & 5.18$\times10^{-5}$ & 5.41$\times10^{-5}$ & 4.28$\times10^{-5}$ & 7.65$\times10^{-5}$ & 4.42$\times10^{-3}$ & 5.75$\times10^{-5}$ & 1.16$\times10^{-4}$ & 1.15$\times10^{-4}$ & 1.54$\times10^{-4}$ & 1.55$\times10^{-4}$ & 1.67$\times10^{-4}$ & 1.76$\times10^{-4}$ & 2.09$\times10^{-4}$ & 2.04$\times10^{-4}$ & 2.45$\times10^{-4}$ & 2.61$\times10^{-4}$ \\ \hline
            
            Opt(w/o noise) \lbrack~\%~\rbrack
            & 1.13$\times10^{-4}$ & 3.03$\times10^{-5}$ & 7.91$\times10^{-6}$ & 1.05$\times10^{-4}$ & 4.44$\times10^{-3}$ & 2.03$\times10^{-5}$ & 9.07$\times10^{-5}$ & 1.20$\times10^{-4}$ & 1.85$\times10^{-4}$ & 2.05$\times10^{-4}$ & 2.32$\times10^{-4}$ & 2.65$\times10^{-4}$ & 3.23$\times10^{-4}$ & 3.42$\times10^{-4}$ & 4.26$\times10^{-4}$ & 5.21$\times10^{-4}$ \\ \hline
            
            FFT(w/ noise) \lbrack~\%~\rbrack
            & 0.326 & 0.378 & 2.33 & 0.184 & 0.106 & 0.300 & 1.47 & 0.0196 & 0.171 & 0.333 & 0.976 & 0.454 & 0.0301 & 0.127 & 0.535 & 0.261 \\ \hline
            
            Opt(w/ noise) \lbrack~\%~\rbrack
            & 0.112 & 0.400 & 1.56 & 0.225 & 0.173 & 0.301 & 0.0295 & 0.0369 & 0.166 & 0.361 & 1.02 & 0.459 & 0.00822 & 0.178 & 0.601 & 0.288 \\ \hline
      \end{tabular}
    }
\end{table*}

\begin{table}[tb]
    \centering
    \caption{Result for circuit~\ref{item:circuit2}, consisting of 16 $\times$ 16 cross-point resistors, under condition~\ref{item:condition1}, where each resistor is $10~\mathrm{k}\Omega$.}
    \label{tab:result_of_exp1_circuit2}
    \setlength{\tabcolsep}{1pt} 
    \begin{tabular}{c|cc|cc}
       \hline
        \multirow{2}{*}{Experiment}
        & \multicolumn{2}{c|}{FFT}
        & \multicolumn{2}{c}{Opt} \\ \cline{2-5}
        & Max. error [\%] & Mean error [\%] & Max. error [\%] & Mean error [\%] \\ \hline
         w/o noise & \textbf{0.00031} & \textbf{0.00015} & 0.00059 & 0.00022 \\
         w/ noise  & 3.63    & 0.40    & \textbf{3.26}    & \textbf{0.32}    \\ \hline
    \end{tabular}
\end{table}

\begin{table}[tb]
    \centering
    \caption{Summary of error rates for each trial}
    \label{tab:error_summary_multi}
    \begin{tabular}{c|cc|cc}
       \hline
        \multirow{2}{*}{Condition}
        & \multicolumn{2}{c|}{FFT}
        & \multicolumn{2}{c}{Opt} \\ \cline{2-5}
        & Mean [\%] & SD [\%] & Mean [\%] & SD [\%] \\ \hline
        \ref{item:condition2} & 0.62 & 4.00 & \textbf{0.37} & \textbf{0.45} \\ \hline
        \ref{item:condition3} & 0.57 & 1.70 & \textbf{0.43} & \textbf{0.38} \\ \hline
    \end{tabular}
\end{table}

\section{Experiments}
We evaluated the  method using the following two circuits:
\begin{enumerate}[label=Circuit~(\arabic*) , align=left, leftmargin=*]
    \item A circuit with 16 parallel resistors \label{item:circuit1}
    \item A $16 \times 16$ cross-point resistor array \label{item:circuit2}
\end{enumerate}

The circuits were simulated in LTspice\cite{analogdevices_ltspice_2026} for 30~ms.
Sinusoidal voltage sources were used as inputs to each channel at 1~kHz intervals from 5~kHz to 20~kHz, with a signal amplitude of 2.5 V and a 2.5 V DC offset.
The current measurement circuit for acquiring $I_m(t)$ consisted of a transimpedance amplifier (TIA), with a $560~\Omega$ feedback resistor.
This is followed by an inverting amplifier for positive-voltage measurement.
The measurement voltage was clipped to the range from 0~V to 5~V and quantized with 16~bit resolution.
Gaussian noise with a standard deviation of 0.01~V was subsequently added to simulate fluctuations observed in practical measurements.
As the baseline method, we employed the AC measurement method  of Hu et al.~\cite{hu2019high}.
For the proposed method, nonlinear least-squares optimization was performed  using the Levenberg--Marquardt algorithm~\cite{marquardt1963algorithm} as a solver.
The initial values were determined from prior knowledge of the applied input signals.
The maximum number of iteration was set to 
1,000,00 as an upper bound.

We evaluated the following three conditions:
\begin{enumerate}[label=(\alph*)]
    \item For principle verification, all resistors in \ref{item:circuit1} and \ref{item:circuit2} were set to $10~\mathrm{k}\Omega$, following a setting similar to that used in the previous study. 
    Both pseudo-ADC without noise and noisy pseudo-ADC conditions were tested to compare the baseline and proposed methods under controlled settings.\label{item:condition1}  
    
    \item In \ref{item:circuit1}, each of the 16 resistors was randomly set in the range from $1\mathrm{~k}\Omega$ to $20\mathrm{~k}\Omega$. 
    This condition was used to evaluate the robustness of resistance measurement accuracy against different resistor combinations. 
    This experiment was repeated 1,000 times.\label{item:condition2}
    
    \item In \ref{item:circuit2}, the resistance distribution was configured to emulate a tactile sensor response. 
    One resistor was randomly selected and assigned a value in the range from $1\mathrm{~k}\Omega$ to $20\mathrm{~k}\Omega$, and the surrounding resistors were varied according to a Gaussian distribution with randomly selected the standard deviation.
    Since the resistance in an active tactile sensor region generally decreases in response to contact, the resistors in inactive resistors were fixed at $20\mathrm{~k}\Omega$.
    This experiment was repeated 1000 times.\label{item:condition3}
\end{enumerate}

\section{Results and Discussions}
Table~\ref{tab:result_of_exp1_circuit1} compares the measurement error rates of the baseline method (FFT) and the proposed method (Opt) for Circuit~\ref{item:circuit1} under condition~\ref{item:condition1}. Without ADC noise, both methods achieved error rates below 0.005~\%. With Gaussian noise, however, the proposed method reduced the maximum and average error rates by 33.0~\% and 26.0~\%, respectively, compared with the baseline method.
Table~\ref{tab:result_of_exp1_circuit2} shows the results for Circuit~\ref{item:circuit2} under condition~\ref{item:condition1}. Although both methods showed comparable accuracy without ADC noise, the error distribution differed across frequencies. Under Gaussian noise, the proposed method reduced the maximum and average error rates by 10.2~\% and 20.0~\%, respectively.
Table~\ref{tab:error_summary_multi} summarizes the mean measurement error rates and standard deviations over 1,000 trials under conditions~\ref{item:condition2} and~\ref{item:condition3}. The proposed method reduced the mean error rates by 40.3~\% and 24.6~\%, respectively, and the standard deviations by 89.0~\% and 77.6~\%, respectively, compared with the baseline method.

The proposed method reduced both the mean error rate and standard deviation over repeated trials under conditions~\ref{item:condition2} and~\ref{item:condition3}, indicating more robust measurement performance under noisy conditions. 
This improvement is likely because the method estimates resistance values by fitting the observed time-series waveform while simultaneously optimizing multiple parameters, which averages out the influence of noise at individual sample points.
Furthermore, the proposed method was more effective than the baseline method not only for a $1 \times N$ array but also for an $N \times M$ cross-point array under load conditions designed to imitate the response of a tactile sensor. 
On the other hand, in the cross-point array, the error distribution differed depending on the frequency even under identical resistance conditions.
When the frequency was fixed, similar error trends were observed for identical resistance conditions, and this tendency was common to both the baseline and proposed methods.
Both the baseline and proposed methods do not consider the sneak path current that is one of the parasitic characteristics of the cross point array circuits.
These results suggest that the relationship between the voltage-source frequency and the parasitic characteristics of the circuit affects the measurement accuracy in AC measurement. 
Therefore, when signals are distinguished using frequency information, the selection of the applied frequencies and circuit models are also important for achieving high measurement accuracy.

\section{Conclusions}
In this study, we proposed a time-domain optimization method for robust readout in AC measurement of cross-point resistive arrays. 
By introducing time-domain waveform fitting into AC measurement, the proposed method improved measurement robustness under noisy observation conditions.
The results showed that the proposed method achieved accuracy comparable to that of the baseline method under noiseless conditions, while reducing the measurement error under noisy conditions.
In particular, the proposed method was also effective for resistance distributions designed to emulate tactile sensor responses, reducing the mean error and standard deviation by 24.6\% and 77.6\%, respectively.

\newpage
\normalsize
\bibliographystyle{IEEEtran}
\bibliography{IEEEabrv,biblio_traps_dynamics}

\end{document}